\documentclass[lettersize,journal]{IEEEtran}
\usepackage{amsmath,amsfonts}
\usepackage{algorithmic}
\usepackage{algorithm}
\usepackage{array}
\usepackage[caption=false,font=normalsize,labelfont=sf,textfont=sf]{subfig}
\usepackage{textcomp}
\usepackage{stfloats}
\usepackage{url}
\usepackage{verbatim}
\usepackage{graphicx}
\usepackage{cite}
\usepackage{booktabs}
\usepackage[table]{xcolor} 
\definecolor{babyblue}{RGB}{137,207,240} 
\definecolor{myblue2}{RGB}{191,239,255}
\usepackage{pifont}
\newcommand{\cmark}{\ding{51}}%
\newcommand{\xmark}{\ding{55}}%
\usepackage{multirow} 
\usepackage{tabularx}
\usepackage[colorlinks=true, urlcolor=blue]{hyperref}

\begin{document}

\title{FADRA: Frequency-Aware Diffusion with Residual Adaptation for Video Face Restoration}

\author{
Jin Jiang\textsuperscript{1} \quad
Jia Wang\textsuperscript{2} \quad
Panwen Hu\textsuperscript{3} \quad
Weiran Zhao\textsuperscript{1} \quad
Shengcai Liao\textsuperscript{1$\dag$}
\\[2mm]
\textsuperscript{1}UAEU \quad
\textsuperscript{2}OUC \quad
\textsuperscript{3}MBZUAI \quad

\thanks{Jin Jiang, Weiran Zhao, and Shengcai Liao are with the College of Information Technology (CIT), United Arab Emirates University (UAEU), Al Ain, Abu Dhabi, United Arab Emirates.}

\thanks{Jia Wang is with the State Key Laboratory of Physical Oceanography and the Faculty of Information Science and Engineering, Ocean University of China (OUC), Shandong, 266100, China.  }

\thanks{Panwen Hu is with the Computer Vision Department, Mohamed bin Zayed University of Artificial Intelligence (MBZUAI), Abu Dhabi, United Arab Emirates.}

\thanks{$\dag$ Corresponding author: Shengcai Liao (email: scliao@ieee.org).}}

\maketitle

\begin{abstract}

Video face restoration (VFR) aims to recover high-quality and temporally consistent facial details from severely degraded video sequences; however, existing methods still struggle to balance spatial fidelity and temporal coherence under complex degradations. To address this, we propose FADRA, a frequency-aware diffusion framework with iterative residual adaptation specifically tailored for robust VFR.
We first leverage the strong temporal consistency of a pre-trained text-to-video diffusion model and introduce lightweight LoRA adapters together with a Low-Quality (LQ) Pixel-Alignment Feature Fusion module to efficiently adapt the frozen generative prior to the VFR task.
To further adapt the frozen diffusion backbone to the downstream VFR task beyond LoRA-based adaptation, we introduce a Repeated Residual Adaptation Head (RRAH) for step-wise residual refinement after the diffusion backbone. To make this refinement explicitly guided by the degraded observation, RRAH further takes the LQ latent together with the current velocity prediction as input, allowing the model to repeatedly revisit LQ cues and predict residual updates at each flow-matching step. This LQ-guided repeated residual adaptation helps recover fine facial details while preserving the inherent temporal priors of the pre-trained model.
Furthermore, to ensure the structural integrity of perceptually important details, we introduce a Frequency-Aware Loss that provides explicit supervision across multiple spectral bands, emphasizing visually sensitive frequency components that are crucial for perceptual quality and prone to temporal jittering.
Extensive experiments demonstrate that FADRA recovers better facial structures and produces more temporally consistent videos than state-of-the-art methods, leading to clear gains in both quantitative metrics and visual perception. Our project page is available at
\href{https://fadra-vfr.github.io/}{https://fadra-vfr.github.io/}.

\end{abstract}


\section{Introduction}
\label{sec:intro}

Video face restoration (VFR) aims to recover high-quality and temporally consistent facial details from severely degraded video sequences. Despite recent progress, VFR remains highly challenging due to the need to simultaneously reconstruct fine-grained facial structures and maintain temporal coherence across frames under complex degradations such as blur, noise, and compression artifacts~\cite{EDVR,BasicVSRpp,VRT,RealESRGAN}. The problem becomes even more difficult when the input videos exhibit large pose variations or expressive facial motions.

\begin{figure}[t]
    \centering
    \includegraphics[width=0.5\textwidth]{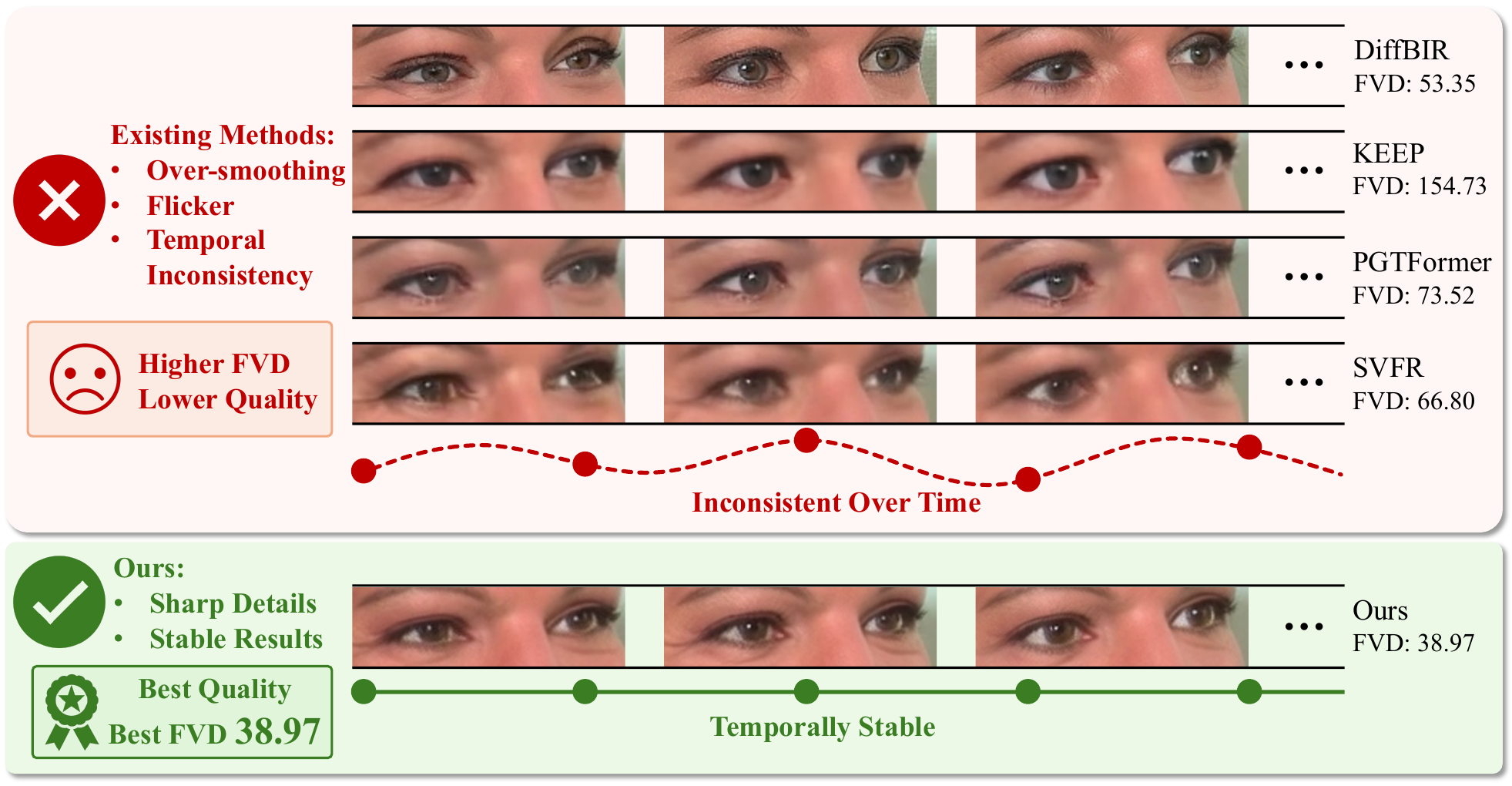} 
   \caption{Visual comparison with state-of-the-art methods on video frames. Existing methods often produce over-smoothed facial details and temporally inconsistent restorations, leading to flickering artifacts and higher FVD scores. In contrast, FADRA restores sharper eye details and maintains a stable appearance across frames, achieving the best FVD on the VFHQ test set.}

    \label{fig:intro}
    \vspace{-3mm}
\end{figure}

Existing VFR approaches~\cite{restoreformer++,keep2024 ,PGTFormer} typically follow either frame-wise restoration or temporal aggregation paradigms. Frame-based methods process each frame independently and thus may achieve reasonable spatial quality, but they often suffer from noticeable temporal flickering and identity drift. In contrast, video-based models aggregate information over time to improve consistency, yet they often oversmooth high-frequency facial details, e.g., eyes, mouth, and teeth, that are crucial for perceptual quality.

Recently, diffusion models have demonstrated remarkable performance on image and video restoration tasks ~\cite{Difface,SVFR,DDRM,DR2} thanks to their strong generative priors and ability to model complex data distributions. However, standard diffusion~\cite{DDPM,LDM} or flow-matching frameworks~\cite{flowmatching} typically predict noise or velocity conditioned on the current state, but lack an explicit mechanism to make full use of the low-quality input cues throughout the trajectory. Once early predictions underutilize or misinterpret these cues, the model may fail to recover subtle facial structures, particularly under severe degradations. Moreover, existing diffusion-based restoration methods~\cite{PFStorer,FaceMe} mainly rely on pixel-space or latent-space losses that implicitly treat different frequency components equally. Such uniform supervision is suboptimal for facial restoration, where high-frequency structures and textures carry important perceptual cues and are especially prone to temporal jittering~\cite{FFL}.


To address the above issues, we propose FADRA, a frequency-aware diffusion framework with iterative residual adaptation specifically tailored for robust VFR. We first leverage the strong temporal consistency inherent in a pre-trained text-to-video diffusion model and introduce lightweight LoRA adapters together with LQ pixel-alignment feature fusion, enabling the frozen generative prior to be efficiently adapted to the VFR task while preserving its temporal modeling capability. Building upon this foundation, one key challenge is how to further adapt the frozen T2V diffusion prior to pixel-aligned restoration. Although lightweight LoRA adapters and the proposed LQ pixel-alignment feature fusion enable efficient task adaptation, recent studies have shown that adapter-based updates may still face challenges in balancing downstream task adaptation and preservation of pre-trained knowledge~\cite{CorDA,OPLoRA,LoRANull}. In our setting, such adaptation may still be insufficient to fully exploit subtle degraded cues throughout the flow-matching trajectory. In particular, after the initial feature fusion, fine-grained facial structures in the LQ observation may still be weakened or overlooked in later prediction steps.

To overcome this limitation, we design a Repeated Residual Adaptation Head (RRAH) after the diffusion backbone. RRAH first provides an additional step-wise residual adaptation pathway beyond LoRA-based backbone adaptation. Furthermore, by taking the LQ latent together with the current velocity prediction as input, RRAH explicitly revisits the degraded cues at each flow-matching step and predicts residual updates for LQ-guided refinement. This design allows FADRA to progressively recover subtle facial details while preserving the strong temporal priors of the pre-trained text-to-video diffusion model.

Furthermore, we introduce a Frequency-Aware Loss (FAL) in the latent frequency domain, which uses a human visual system (HVS)-inspired luminance prior to reweight spectral components. This provides explicit supervision across multiple spectral bands and better preserves perceptually important facial structures.

Our contributions can be summarized as follows:

1. We propose FADRA, a novel diffusion-based VFR framework that adapts a frozen text-to-video diffusion backbone with lightweight LoRA modules and low-quality guidance, enabling high-fidelity and temporally stable restoration.

2. A repeated residual adaptation head is designed to complement LoRA-based adaptation with LQ-guided step-wise residual refinement, allowing the model to ``re-examine'' degraded cues for identity-preserving detail recovery.

3. A frequency-aware loss based on an HVS-inspired luminance prior is introduced to explicitly supervise latent frequency components and better preserve perceptually important facial structures.

\vspace{-3mm}

\section{Related Work}

\subsection{Image Face Restoration}

Image Face Restoration (IFR) aims to recover facial details from low-quality face images~\cite{DFDNet,CTCNet-TIP,SGPN,UMSN-TIP}. Early IFR methods~\cite{GFPGAN,GPEN,SCGAN-TIP} apply GAN-based architectures for face restoration, recovering notable facial details and global appearance. Subsequent Transformer-based approaches, such as CodeFormer~\cite{CodeFormer}, RestoreFormer~\cite{restoreformer}, and DAEFR~\cite{DAEFR}, further exploit long-range dependencies and rich latent priors to explicitly model facial structure and degradation, achieving a more favorable balance between perceptual realism and identity fidelity. In recent years, driven by the impressive image generative capabilities of diffusion models, advanced approaches such as PGDiff~\cite{PGDiff}, DifFace~\cite{Difface}, DiffBIR~\cite{DiffBIR}, and OSDFace~\cite{OSDFace} leverage powerful pre-trained diffusion priors to recover realistic facial details under unknown degradations. They often combine degradation-aware conditioning or reference guidance to improve robustness, achieving state-of-the-art performance on challenging blind face restoration benchmarks. However, they suffer from noticeable temporal inconsistency when extended to videos in a frame-by-frame manner. In contrast, FADRA is designed directly for videos, leveraging a text-to-video diffusion prior and explicit temporal modeling to jointly improve spatial fidelity and temporal coherence.

\begin{figure*}[t]
    \centering
    \includegraphics[width=1\textwidth]{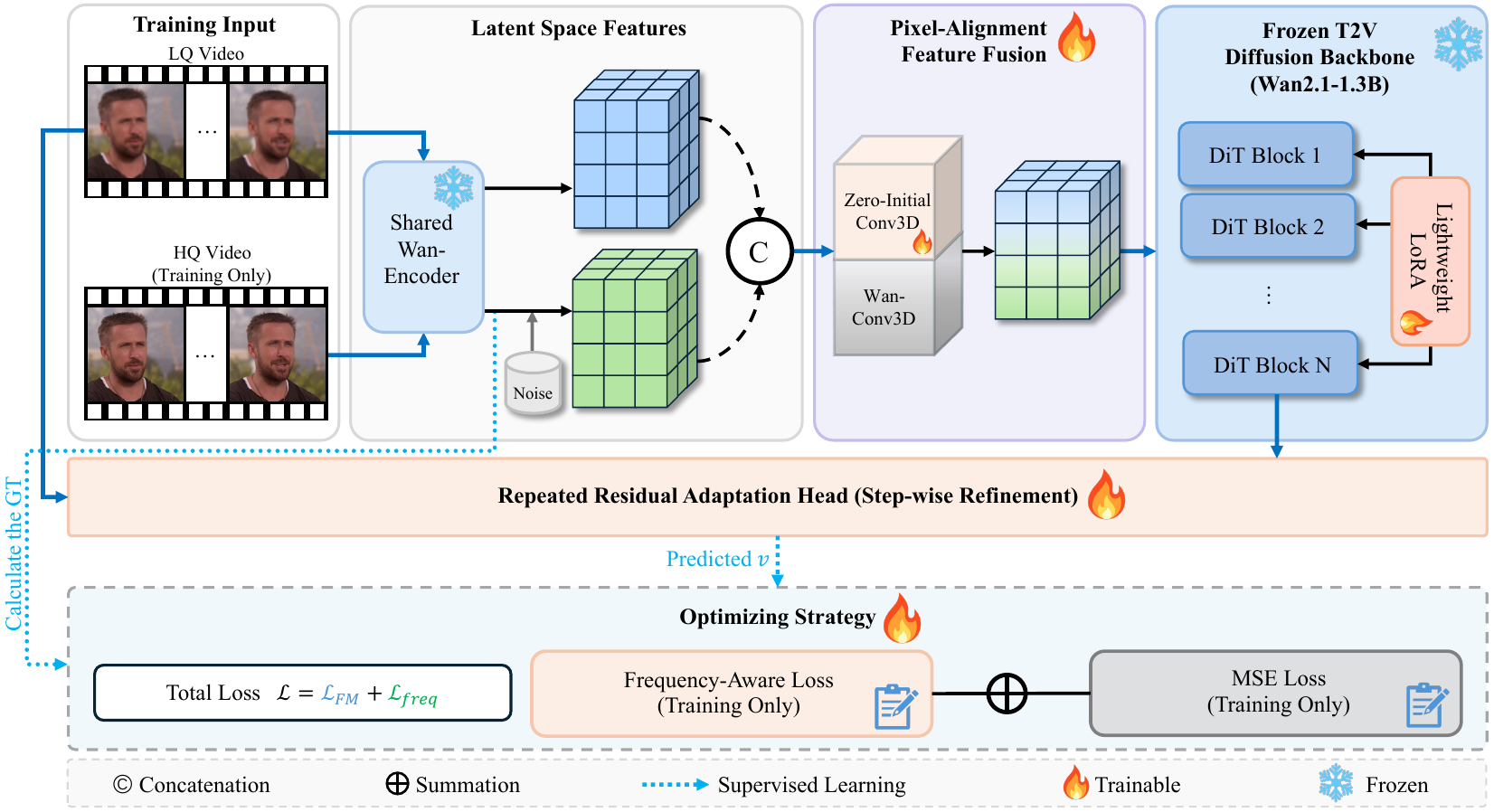} 
    \caption{Overview of the proposed FADRA framework. During training, low-quality (LQ) and high-quality (HQ) videos are both encoded into the latent space to extract spatio-temporal features. We adapt a frozen diffusion backbone using lightweight LoRA modules to align it with the specific requirements of the VFR task. A Repeated Residual Adaptation Head (RRAH) is integrated to refine velocity predictions at each flow-matching step. A Frequency-Aware Loss (FAL) provides explicit spectral supervision on the predicted latents to enhance perceptual fidelity. }
    
    \label{fig:framework}
    \vspace{-3mm}
\end{figure*}

\subsection{Video Face Restoration}
Compared with image-based settings, video face restoration (VFR) is more challenging due to the need to balance spatial fidelity and temporal coherence under complex degradations. Existing real-world video super-resolution (VSR) methods~\cite{RealBasicVSR,StableVSR,MGLDVSR,VRT-TIP} and diffusion-based video enhancement methods~\cite{STAR,VEnhancer} have achieved notable progress on generic natural scenes and can yield moderate improvements on face-containing videos. However, as generic frameworks, these methods do not explicitly exploit facial structure or identity priors and still struggle under the severe degradations that are common in in-the-wild face videos. To better leverage facial structures and temporal cues, several works design specialized VFR architectures. Earlier works, such as KEEP~\cite{keep2024} and PGTFormer~\cite{PGTFormer}, attempt to enhance temporal consistency by maintaining recurrent face priors or designing the temporal-coherent Transformer architectures. TCN~\cite{TCN-TIP} is proposed to facilitate the extension of blind face image restoration methods to videos by enhancing temporal consistency with alignment smoothing strategy, thereby reducing facial jitters and flickering artifacts. More recently, SVFR~\cite{SVFR} generalizes video face restoration into a unified framework by conditioning Stable Video Diffusion~\cite{SVD} on task-specific embeddings, achieving improved temporal coherence across heterogeneous tasks. Nevertheless, the temporal stability of these approaches remains limited when facing long sequences or complex, real-world degradations.

In contrast, FADRA leverages the strong temporal-consistency priors of the DiT-based text-to-video diffusion model and introduces a repeated residual adaptation head and a frequency-aware loss as two key components specifically designed for VFR. These designs complement existing VFR methods and make our framework particularly effective for restoring high-quality, temporally stable face videos under complex real-world degradations.

\section{Method}

\subsection{Overall Framework}
\label{sec:framework}

As illustrated in Fig.~\ref{fig:framework}, FADRA builds upon the powerful text-to-video diffusion model Wan~\cite{wan2025} and adapts it to the video face restoration task in a parameter-efficient manner. Given a low-quality (LQ) input video and its corresponding high-quality (HQ) ground truth during training, both sequences are encoded by the frozen Wan VAE encoder into spatio-temporal latent representations. The pretrained Wan backbone is kept frozen, and we introduce lightweight LoRA adapters~\cite{lora} into the DiT backbone to better align the latent space with the VFR task while preserving the strong generative and temporal priors of the original model.

To inject degraded video information for pixel-aligned restoration, the encoded LQ latent and the noised HQ latent are first concatenated along the channel dimension and fed into an LQ Pixel-Alignment Feature Fusion module. Specifically, we expand the input channels of the original Wan 3D convolution layer to accommodate the additional LQ guidance. The newly added base weights are initialized to zero, ensuring that the expanded model initially preserves the behavior of the pretrained Wan backbone while enabling stable and efficient LQ injection. LoRA adapters are then applied to the expanded 3D projection, allowing the model to learn a residual adaptation that jointly calibrates the noisy latent stream and the LQ guidance stream. The fused representation is then projected back to the original input dimension of the DiT backbone and subsequently processed by the frozen Wan backbone with trainable adapters. In this way, the Wan backbone can already be adapted to VFR without introducing heavy auxiliary conditioning branches.

To further adapt the frozen diffusion backbone beyond LoRA-based adaptation, we introduce a Repeated Residual Adaptation Head (RRAH) after the DiT backbone. RRAH provides a lightweight step-wise residual refinement pathway for the velocity prediction. To make this refinement explicitly guided by the degraded observation, RRAH takes the current velocity prediction concatenated with the LQ latent as input and predicts a residual update, which is then added back to obtain a refined prediction $\hat{v}_{t}$. During inference, this residual refinement is applied at each of the $T$ sampling steps, forming an LQ-guided repeated residual adaptation scheme. In this way, the Wan backbone provides strong motion and content priors, while RRAH specializes the model for VFR by more effectively exploiting LQ cues. 

The entire framework is optimized in a supervised fashion with two complementary objectives, including a standard MSE loss and the proposed frequency-aware loss. Both losses are used only during training. The MSE loss encourages overall fidelity to the ground truth, whereas the frequency-aware loss explicitly reweights spectral components according to an HVS-inspired frequency prior, helping preserve perceptually important facial structures. During inference, FADRA requires only the LQ video and noise as input, performs $T$-step refinement in the latent space, and then uses a single pass of the Wan decoder to obtain the final restored video.

\subsection{Repeated Residual Adaptation Head}

\begin{figure*}[t]
    \centering
    \includegraphics[width=1\textwidth]{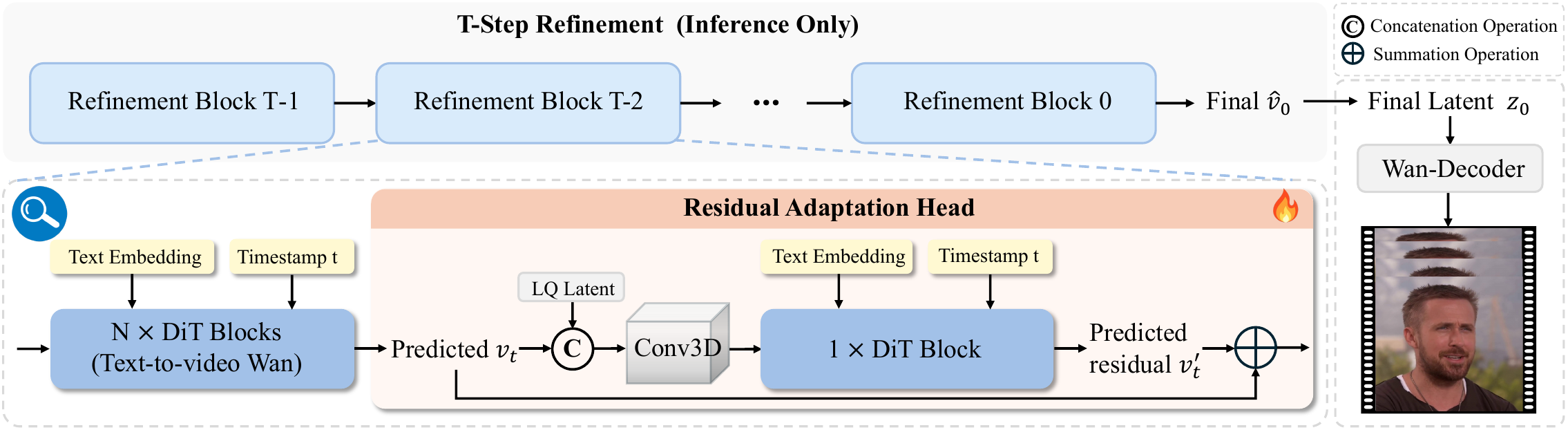} 
    \caption{Illustration of the LQ-guided ``re-examination'' mechanism in the Repeated Residual Adaptation Head (RRAH).}
    \label{fig:PRCH}
    \vspace{-3mm}
\end{figure*}

Video face restoration requires strict spatial alignment with the LQ input to ensure identity fidelity. However, text-to-video (T2V) diffusion models~\cite{wan2025,cogvideo,cogvideox,sanavideo} are primarily optimized for high-quality generation and temporal motion consistency, rather than precise pixel-aligned restoration. This discrepancy between generative priors and restoration requirements often leads to a ``fidelity gap''. Although LoRA-based adaptation and LQ pixel-alignment feature fusion can efficiently adapt the frozen backbone to VFR, they may still be insufficient to fully exploit subtle degraded cues throughout the flow-matching trajectory.

To address this issue, we propose the Repeated Residual Adaptation Head (RRAH) as a lightweight module complementary to LoRA-based adaptation. In the context of VFR, the diffusion backbone provides a coarse generative trajectory, while RRAH performs step-wise residual refinement after the backbone to improve high-fidelity facial reconstruction without compromising the temporal priors of the pre-trained model. More importantly, RRAH takes the LQ latent together with the current velocity prediction as input and predicts a residual update at each flow-matching step, allowing the model to iteratively revisit degraded cues and strengthen the alignment between the T2V generative prior and the high-fidelity requirements of the VFR task.

As illustrated in Fig.~\ref{fig:PRCH}, the diffusion backbone generates an initial velocity prediction ${v}_t$ at each flow-matching step $t$. To enhance LQ guidance, the RRAH takes the current velocity prediction ${v}_t$ and the original LQ latent ${z}_{lq}$ as inputs. We first fuse ${v}_t$ and ${z}_{lq}$ through a concatenation operation followed by a 3D convolutional layer for dimensional alignment:\begin{equation}{F}_{fuse} = \text{Conv3D}(\text{Concat}({v}_t, {z}_{lq})).
\end{equation}
The fused features ${F}_{fuse}$ are then processed by a lightweight DiT block to predict a residual velocity ${v}'_t$. The final velocity field can be formulated as a residual update as follows:
\begin{equation}
\hat{v}_{t} = {v}_t + {v}'_t.
\end{equation}
By iteratively ``re-examining'' the LQ latent throughout the $T$-step refinement, the RRAH helps to capture fine-grained facial details that are often overlooked by the primary backbone. 
Furthermore, since the RRAH predicts residual updates ${v}'_t$ rather than replacing the velocity ${v}_t$ generated by the diffusion backbone, the backbone still governs the primary temporal motion and structural consistency of the pre-trained model, while the RRAH provides targeted residual adaptation and fidelity alignment.

\subsection{Frequency-Aware Loss}

Facial perception in videos is highly sensitive to high-frequency structures such as eyes, mouth, and fine textures, whose temporal inconsistency often leads to noticeable flickering artifacts. However, conventional text-to-video diffusion models are typically trained with flow-matching and MSE losses applied directly to the predicted velocity 
$v$, treating all frequency components equally, which is suboptimal for video face restoration. To address this, we introduce a Frequency-Aware Loss that explicitly supervises different frequency components of the video latent and better preserves perceptually important details. Our design is inspired by the frequency-decoupled objective in DeCo~\cite{deco}, but extends it from image generation to VFR and from RGB domain to multi-channel diffusion latents.

Unlike DeCo, which performs frequency decoupling on the pixel-space velocity $v_t$ used for flow matching, we first reconstruct the corresponding latent representation $z_t$ from $v_t$ and apply the discrete cosine transform (DCT) in the latent space, since the spectral content of these latents is more tightly coupled to the underlying facial structures, where their spectral coefficients typically correspond to spatial variations in the facial content. As a result, our frequency-aware supervision operates on a representation where the frequency components have a more direct and physically interpretable connection to facial details.

Concretely, given the predicted and ground-truth latents $\hat{z}_t$ and $z_t$, 
we transform the latents from the spatial domain to the frequency domain using a block-wise $8\times 8$ DCT. For video inputs, we first reshape the temporal dimension into the batch dimension,  then apply the DCT to each frame, and finally reshape back to recover the original spatio-temporal structure. This transform can be denoted by $\mathcal{T}(\cdot)$ as follows:
\begin{equation}
    \hat{Z}_t = \mathcal{T}(\hat{z}_t), \qquad
    Z_t = \mathcal{T}(z_t).
\end{equation}

Given the key observation that the human eye has different sensitivities to luminance changes and different frequency components, we emphasize perceptually important frequencies for Frequency-Aware Loss following the JPEG quantization strategy, which is designed according to properties of the human visual system (HVS)~\cite{hvs}.

JPEG defines an $8\times 8$ quantization table for the luminance component.
Following the JPEG specification~\cite{JPEG}, we first obtain a scaled table $Q_{\text{cur}}$ for a given quality factor $q\in[1,100]$ as in DeCo~\cite{deco}:
\begin{equation}
    Q_{\text{cur}} = 
    \max\!\left(
        1,\;
        \left\lfloor
            \frac{Q_{\text{base}}\cdot (100-q) + 25}{50}
        \right\rfloor
    \right),
    \label{eq:q_cur}
\end{equation}
where $Q_{\text{base}} \in \mathbb{R}^{8\times 8}$ denotes the standard base luminance table. We then convert $Q_{\text{cur}}$ into frequency weight by taking a normalized inverse:
\begin{equation}
    \tilde{w} = \left(\frac{\bar{Q}_{\text{cur}}}{Q_{\text{cur}}}\right)^{\gamma}, 
    \qquad
    w = \frac{\tilde{w}}{\mathbb{E}[\tilde{w}]},
    \label{eq:w_from_q}
\end{equation}
where $\bar{Q}_{\text{cur}}$ is the mean of $Q_{\text{cur}}$ and $\gamma \ge 1$ controls the strength of frequency reweighting.

In our setting with multi-channel latent representations, we similarly replicate the luminance weights across all channels, yielding
$w \in \mathbb{R}^{C\times 8\times 8}$. This design effectively encodes the HVS-inspired luminance prior while remaining agnostic to the semantic meaning of each latent channel.

The frequency-aware loss is then defined as a weighted regression in the frequency domain:
\begin{equation}
    \mathcal{L}_{\text{freq}}
    = \mathbb{E}_{x,t} \Big[\, 
        w \,\big\| \hat{Z}_t - Z_t \big\|_2^2
    \Big].
    \label{eq:freq_loss}
\end{equation}
By applying this normalized frequency weighting, $\mathcal{L}_{\text{freq}}$ encourages the model to allocate more capacity to reconstruct fine facial structures, rather than treating all content equally. Since the loss is evaluated on all frames of the sequence, it provides consistent supervision across time and helps reduce temporal flickering in visually sensitive regions.

Finally, the overall training objective is formulated as the sum of the standard flow-matching loss $\mathcal{L}_{\text{FM}}$ and the proposed frequency-aware loss:
\begin{equation}
    \mathcal{L}
    = \mathcal{L}_{\text{FM}} 
    + \mathcal{L}_{\text{freq}},
    \label{eq:total_loss}
\end{equation}
where the flow-matching loss $\mathcal{L}_{\text{FM}}$ is defined as an MSE loss between the predicted and target velocities.

\section{Experiments}
\subsection{Datasets and Implementation Details}
\subsubsection{Datasets and Training Details} 
To ensure that the model can handle diverse facial identities and motions, we employ a hybrid dataset derived from VFHQ~\cite{VFHQ} and HDTF~\cite{HDTF}. The training set is a refined subset consisting of 15,127 high-quality video clips. For evaluation, we follow the official VFHQ test split, which consists of 50 representative sequences, ensuring no identity overlap with the training data. To further assess effectiveness and generalization ability, we also construct an additional test set by randomly sampling 20 high-quality video sequences from CelebV-HQ~\cite{CelebvHQ}. All methods are evaluated on this test set without additional training or hyper-parameter tuning on CelebV-HQ.

Our framework is implemented in PyTorch and we adopt the Wan2.1-T2V-1.3B~\cite{wan2025} as the pre-trained foundation model, leveraging its robust generative priors for temporal modeling. The model is trained on two NVIDIA A100 GPUs for 10,000 iterations with a batch size of 32 and the learning rate is set to $1 \times 10^{-4}$. The LoRA rank is $128$.

\subsubsection{Dataset Degradation Pipeline}
\label{sec:D1} 
To train our model under diverse and realistic distortions, we generate paired low-quality (LQ) videos from high-quality (HQ) face videos using a stochastic degradation pipeline. 
Given an input clip $\mathbf{x}\in\mathbb{R}^{F\times3\times H\times W}$, we apply a sequence of randomly sampled operations that simulate typical real-world degradations, including resolution changes, blur, noise, and compression. 
All degradation parameters are shared across frames within a clip to preserve temporal coherence.

Specifically, we first apply resolution changes and blur to the video by randomly sampling a resize ratio $r \sim \mathcal{U}(1.0, 8.0)$ and computing the degraded spatial size as $\hat{H} = \mathrm{round}(H / r)$ and $\hat{W} = \mathrm{round}(W / r)$. Each frame is first downsampled to $(\hat{H}, \hat{W})$ and then upsampled back to $(H, W)$ using bilinear interpolation, which introduces typical resampling artifacts such as aliasing and interpolation blur. We further apply the Gaussian blur with a randomly selected odd kernel size $k \in \{3, 5, \dots, 25\}$ and blur strength $\sigma_b \sim \mathcal{U}(1.5, 6.0)$. The overall operation can be written as:

\[
\mathbf{x}' \gets \mathrm{Blur}_k^{\sigma_b}(\mathrm{Resize}_{H,W}(\mathrm{Resize}_{\hat{H},\hat{W}}(\mathbf{x}))).
\]

Then, the blurred frames are normalized to the range $[0,1]$ and perturbed by additive Gaussian noise
$\mathcal{N}(0,\sigma_n^2)$ with standard deviation
$\sigma_n\sim\mathcal{U}(0.01, 0.05)$, followed by clipping back to $[0,1]$ to avoid overflow:
\[
\tilde{\mathbf{x}} \gets \mathrm{clip}\bigl(\mathbf{x}' + \mathcal{N}(0,\sigma_n^2)\bigr).
\]

To simulate codec-like compression and transmission artifacts, we further apply a frame-wise JPEG compression operator to the degraded frames in the pixel space.
Concretely, the frames are passed through a random JPEG compression operator with a quality factor
$q \sim \mathcal{U}(25, 85)$ as follows:
\[
\hat{\mathbf{x}} \gets \mathrm{JPEG}_q(\tilde{\mathbf{x}}),
\]
which introduces typical blocking and ringing artifacts commonly observed in compressed web videos. Finally, the resulting clip is used as the LQ input, while the original HQ clip serves as the supervision target during training.

\subsubsection{Evaluation Metrics}
\label{sec:metrics}

We employ both pixel-level, perception-based, and video-level metrics to quantitatively evaluate the performance of FADRA and competing methods.  Concretely, we report PSNR and SSIM~\cite{SSIM} to measure pixel-level fidelity, and LPIPS~\cite{LPIPS} to evaluate perceptual quality. In addition, we calculate the Identity Distance (IDD) by extracting feature embeddings using a pre-trained ArcFace~\cite{ArcFace} and computing the angular distance between the restored frames and their corresponding ground truth, thereby reflecting identity preservation. All these metrics are computed frame-wise and subsequently averaged over all sequences in the test set. Beyond frame-level evaluation, we further adopt the Fréchet Video Distance (FVD)~\cite{FVD} as a video-level metric to measure temporal coherence and overall video realism.  Moreover, we report frames per second (FPS) to quantify the runtime efficiency of different methods.

\subsection{Comparison with State-of-the-Art}

\begin{table*}[t]
\centering
\caption{Overall comparison with state-of-the-art methods on the VFHQ and CelebV-HQ test sets. The best performance is highlighted in bold, while the second-best result is indicated with an underscore. }
\label{tab:sota}
\resizebox{\textwidth}{!}{
\begin{tabular}{lccll}
\toprule
\multirow{2}{*}{\textbf{Method}}
& \multirow{2}{*}{\textbf{Type}}
& \multirow{2}{*}{\textbf{Level}}
& \multicolumn{1}{c}{\textbf{VFHQ}}
& \multicolumn{1}{c}{\textbf{CelebV-HQ}} \\
& & 
& \textbf{PSNR}$\uparrow$ / \textbf{SSIM}$\uparrow$ / \textbf{LPIPS}$\downarrow$ / \textbf{IDD}$\downarrow$ / \textbf{FVD}$\downarrow$
& \textbf{PSNR}$\uparrow$ / \textbf{SSIM}$\uparrow$ / \textbf{LPIPS}$\downarrow$ / \textbf{IDD}$\downarrow$ / \textbf{FVD}$\downarrow$ \\
\midrule
{CodeFormer~\cite{CodeFormer}} 
& Transformer 
& Image 
& 26.93 / 0.7810 / 0.2826 / 0.5173 / 177.11 
& 26.44 / 0.7674 / 0.2977 / 0.5262 / 251.00
\\

{RestoreFormer++~\cite{restoreformer++}}
& Transformer 
& Image 
& 27.28 / 0.7921 / 0.2657 / 0.3875 / 81.14 
& 26.69 / 0.7708 / 0.2807 /	0.3815 /  118.90
\\

DAEFR~\cite{DAEFR}
& Transformer 
& Image 
& 22.79 / 0.6834 / 0.3485 / 0.8044 / 819.31 
& 22.25 / 0.6609 / 0.3591 / 0.7742 / 1005.01
\\

PGDiff~\cite{PGDiff} 
& Diffusion 
& Image 
& 22.35 / 0.6712 / 0.4583 / 1.1174 / 828.43 
& 22.82 / 0.6731 / 0.4558 / 0.9835 / 676.30
\\

DifFace~\cite{Difface} 
& Diffusion 
& Image 
& 25.49 / 0.7315 / 0.3780 / 0.8401 / 2399.60 
& 25.29	/ 0.7253 / 0.3784 / 0.7484 / 2456.44
\\

DiffBIR~\cite{DiffBIR}
& Diffusion 
& Image 
& 27.56 / 0.7678 / 0.2852 / \underline{0.3569} / \underline{53.35} 
& 26.83 / 0.7499 / 0.3030 / 0.3843 / 125.31
\\

OSDFace~\cite{OSDFace}
& Diffusion 
& Image 
& 24.56 / 0.7385 / 0.2948 / 0.4434 / 155.98 
& 24.26	/ 0.7228 / 0.3288 / 0.5020 / 452.52
\\

\midrule

RealBasicVSR~\cite{RealBasicVSR}
& CNN
& Video 
& \underline{28.51} / 0.8160 / 0.3400 / 0.6704 / 172.09 
& 27.66 / 0.7988 / 0.3536 / 0.6308 / 297.70
\\

StableVSR~\cite{StableVSR}
& Diffusion 
& Video 
& 27.79 / 0.8206 / 0.3201 / 0.6909 / 172.03 
& 27.02 / 0.7891 / 0.3730 / 0.6794 / 335.82
\\

MGLDVSR~\cite{MGLDVSR}
& Diffusion 
& Video 
& 28.30 / \underline{0.8271} / 0.3039 / 0.5232 / 109.20 
& 27.30 / 0.7829 / 0.3766 / 0.6181 / 291.73
\\

VEnhancer~\cite{VEnhancer}
& Diffusion 
& Video 
& 20.47 / 0.6903 / 0.4673 / 0.9289 / 1022.32 
& 19.78 / 0.6466 / 0.4784 / 0.9265 / 1538.55
\\

STAR~\cite{STAR}
& Diffusion 
& Video 
& 26.16 / 0.8094 / 0.3497 / 0.4958 / 155.32 
& 25.26 / 0.7798 / 0.3968 / 0.6683 / 367.40
\\

\midrule

KEEP~\cite{keep2024}
& Transformer 
& Video 
& 27.55 / 0.7876 / 0.2708 / 0.5068 / 154.73 
& 26.30 / 0.7611 / 0.3056 / 0.6141 / 510.82
\\

PGTFormer~\cite{PGTFormer} 
& Transformer 
& Video 
& 25.73 / 0.7853 / \underline{0.2610} / 0.3902 / 73.52 
& 25.06 / 0.7646 / 0.2885 / 0.4439 / 149.25
\\

SVFR~\cite{SVFR}
& Diffusion 
& Video 
& 26.54 / 0.7891 / 0.2815 / 0.3782 / 66.80 
& \underline{28.20} / \underline{0.8190} / \underline{0.2513} / \underline{0.3545} / \underline{52.33}
\\

\rowcolor{myblue2!30} 
\textbf{Ours} 
& Diffusion 
& Video 
& \textbf{29.95} / \textbf{0.8576} / \textbf{0.2378} / \textbf{0.2912} / \textbf{38.97} 
& \textbf{28.70} / \textbf{0.8326} / \textbf{0.2421} / \textbf{0.3103} / \textbf{48.32}
\\
\bottomrule
\end{tabular}
}
\vspace{-1mm}
\end{table*}

\begin{figure*}[t]
    \centering
    \includegraphics[width=1\textwidth]{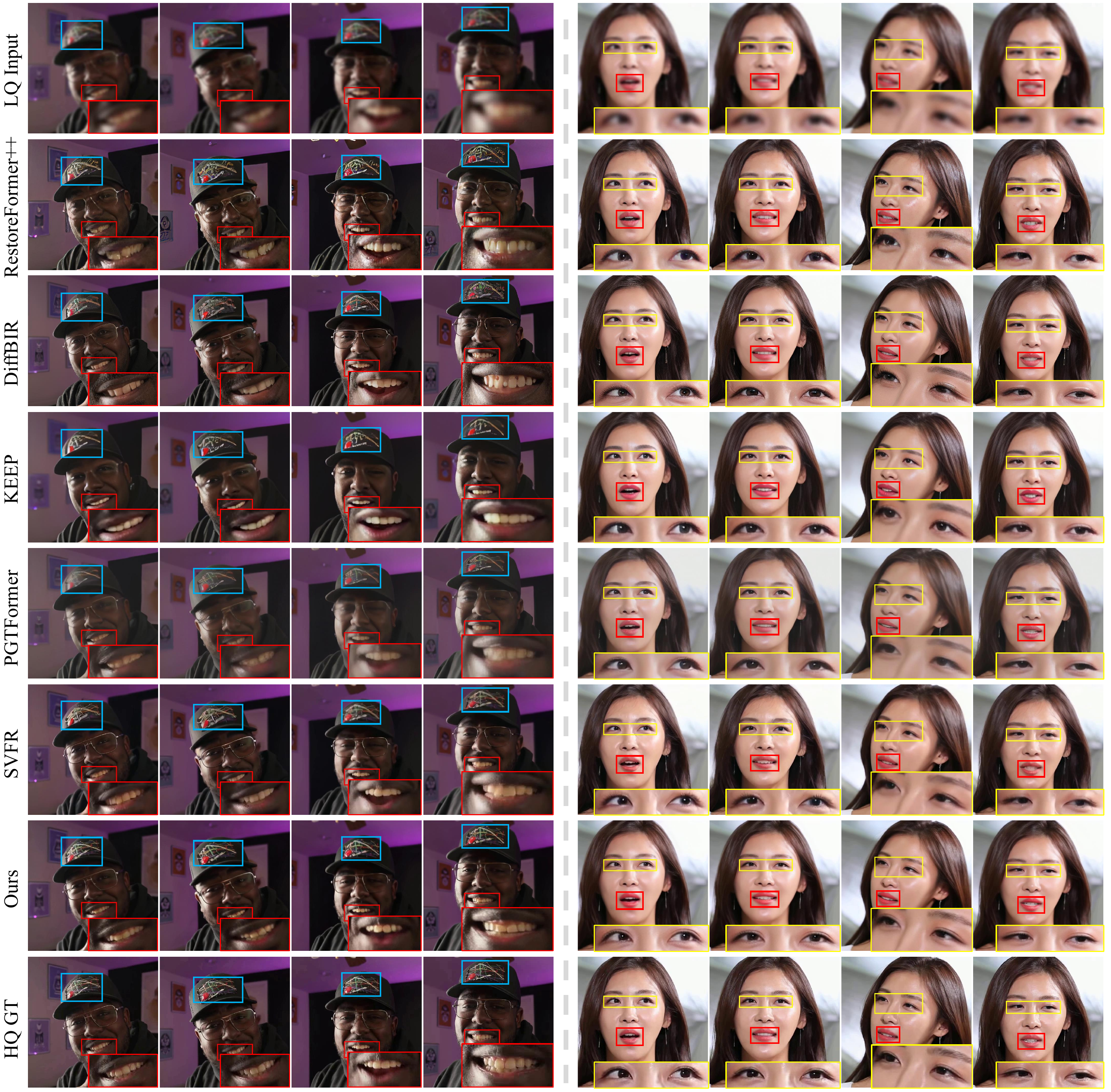} 
\caption{Visual comparison of restoration results from different methods on the VFHQ test set. For two image-based methods, RestoreFormer++~\cite{restoreformer++} and DiffBIR~\cite{DiffBIR}, we apply them frame-by-frame to the degraded videos. To facilitate a clearer comparison of local details, we highlight the eye region with yellow boxes, the mouth and teeth region with red boxes, and other non-facial regions, such as hat textures, with blue boxes. }
    \label{fig:visual_compare}
\vspace{-3mm}
\end{figure*}

\begin{figure*}[t]
    \centering
    \includegraphics[width=0.98\textwidth]{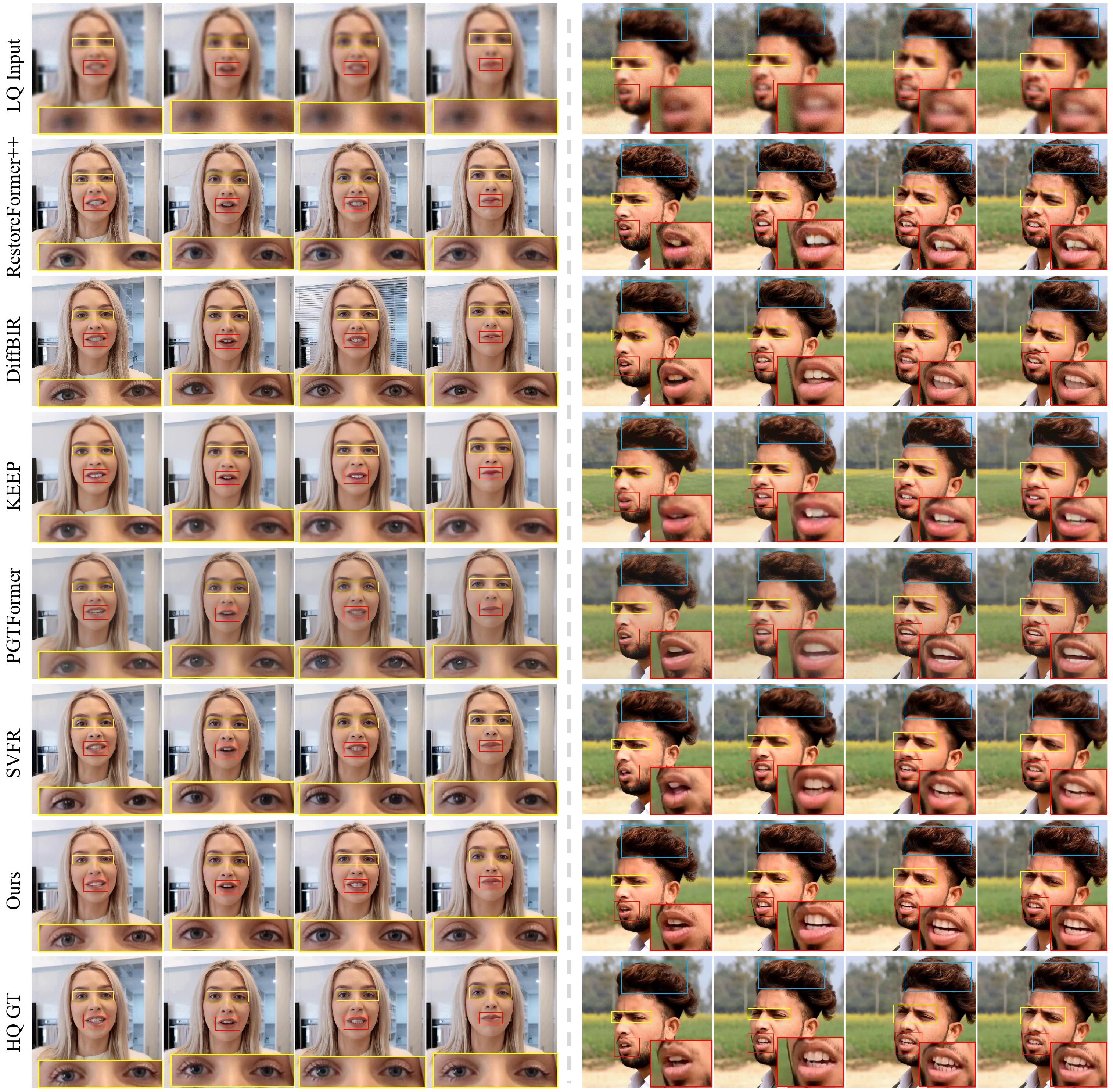} 
\caption{Visual comparison of restoration results from different methods on examples from CelebV-HQ test set. For two image-based methods, RestoreFormer++ and DiffBIR, we apply them frame-by-frame to the degraded videos. To facilitate a clearer comparison of local details, we highlight the eye region with yellow boxes, the mouth and teeth region with red boxes, and hair textures with blue boxes.
}
    \label{fig:visual_compare_celebv-1}
    \vspace{-4mm}
\end{figure*}

\begin{table*}[t]
\centering
\caption{FPS and running time comparison with state-of-the-art diffusion methods for a 161-frame $512 \times 512$ video generation on a single NVIDIA A100 GPU, with the best performance in bold and the second best underlined.}
\label{tab:FPS_compare}
\resizebox{\textwidth}{!}{
\begin{tabular}{lcccccccccc}
\toprule
\textbf{Method}
& \textbf{{PGDiff} }
& \textbf{{DifFace} }
& \textbf{{DiffBIR} }
& \textbf{{OSDFace} }
& \textbf{{StableVSR} }
& \textbf{{MGLDVSR} }
& \textbf{{VEnhancer}}
& \textbf{{STAR}}
& \textbf{{SVFR}}
& \textbf{{Ours}} \\
\midrule
\textbf{FPS $\uparrow$} 
& 0.014
& 0.265
& 0.148
& \textbf{1.988}
& 0.129
& 0.013
& 0.168
& 0.172
& 0.770
& \underline{0.866} \\
\textbf{Time (s) $\downarrow$} 
& 11372
& 608
& 1089
& \textbf{81}
& 1248
& 12716
& 960
& 935
& 209
& \underline{186}\\
\bottomrule
\end{tabular}
}
\vspace{-3mm}
\end{table*}

\subsubsection{Quantitative Evaluation}

As shown in Table~\ref{tab:sota}, we compare FADRA with representative image-based face restoration methods and video-based restoration methods on the VFHQ and CelebV-HQ test sets. The evaluation covers both frame-level reconstruction quality and video-level temporal consistency, including PSNR, SSIM, LPIPS, IDD, and FVD.

On the VFHQ test set, FADRA achieves the best performance across all five metrics. Specifically, it obtains 29.95 PSNR and 0.8576 SSIM, demonstrating superior pixel-level reconstruction and structural preservation compared with existing image-based and video-based methods. In terms of perceptual quality and identity preservation, FADRA further reduces LPIPS and IDD to 0.2378 and 0.2912, respectively, outperforming the strong diffusion-based VFR baseline SVFR. More importantly, FADRA achieves an FVD of 38.97, substantially improving over SVFR and DiffBIR, whose FVD scores are 66.80 and 53.35, respectively. This indicates that FADRA not only restores more faithful facial details but also produces more temporally coherent and stable video sequences.

On the CelebV-HQ test set, FADRA also consistently outperforms all competing methods, demonstrating strong cross-dataset generalization without any fine-tuning on CelebV-HQ. Among image-based methods, DiffBIR and RestoreFormer++ achieve competitive results, yet they remain noticeably behind FADRA across all metrics. For example, FADRA obtains 28.70 PSNR, 0.3103 IDD, and 48.32 FVD, while DiffBIR reaches 26.83 PSNR, 0.3843 IDD, and 125.31 FVD, and RestoreFormer++ obtains 26.69 PSNR, 0.3815 IDD, and 118.90 FVD. Compared with the strongest video-based baseline SVFR, FADRA further improves PSNR and SSIM from 28.20 and 0.8190 to 28.70 and 0.8326, while reducing LPIPS, IDD, and FVD from 0.2513, 0.3545, and 52.33 to 0.2421, 0.3103, and 48.32, respectively. These results confirm that the proposed frequency-aware diffusion framework with residual adaptation generalizes well to unseen high-quality face videos while preserving identity and temporal coherence.

\subsubsection{Efficiency Analysis} 
Since conventional CNN- and Transformer-based methods show limited competitiveness in Table~\ref{tab:sota}, we focus our efficiency analysis on recent state-of-the-art diffusion-based approaches. As reported in Table~\ref{tab:FPS_compare}, FADRA achieves an inference speed of 0.866 FPS, which is significantly faster than video diffusion models such as StableVSR and STAR, whose speeds are 0.129 FPS and 0.172 FPS, respectively. Compared with the state-of-the-art VFR method SVFR, FADRA not only achieves better restoration quality as shown in Table~\ref{tab:sota}, but also delivers higher inference efficiency, improving the FPS from 0.770 to 0.866 and reducing the processing time from 209s to 186s. Although OSDFace achieves the highest FPS due to its one-step generation paradigm, it is an image-based diffusion model and struggles to maintain temporal consistency across frames. This is reflected by its much higher FVD score of 155.98, compared with 38.97 achieved by FADRA. These results demonstrate that FADRA provides a favorable trade-off between high-fidelity restoration and inference efficiency among existing diffusion-based video face restoration frameworks.

\subsubsection{Qualitative Evaluation}

To visually evaluate the restoration quality and generalization ability of FADRA, we provide qualitative comparisons with state-of-the-art methods on the VFHQ and CelebV-HQ test sets in Fig.~\ref{fig:visual_compare} and Fig.~\ref{fig:visual_compare_celebv-1}, respectively. 

On the VFHQ test set, existing image-based and video-based methods often produce blurry results, introduce artifacts, or distort fine facial structures. In particular, the eye and mouth regions are frequently over-smoothed, leading to unclear ocular structures and unnatural tooth textures. In contrast, FADRA restores sharper and more realistic facial details across frames, including clearer eye structures and more authentic dental textures. For non-facial regions marked with the blue boxes, our method also reconstructs richer patterns on hats while maintaining smooth temporal transitions without noticeable jitter. These results demonstrate that FADRA not only improves the fidelity of primary facial components but also handles diverse surrounding textures in a temporally coherent manner, which is consistent with the quantitative gains reported in Table~\ref{tab:sota}.

On the CelebV-HQ test set, the visual comparisons further validate the generalization ability of FADRA under a zero-shot setting. As shown in Fig.~\ref{fig:visual_compare_celebv-1}, competing methods such as RestoreFormer++, KEEP, and PGTFormer still struggle to recover intricate facial details under severe degradations, especially in the eyes and teeth regions. Some methods such as SVFR and DiffBIR also fail to reconstruct high-frequency hair textures, as highlighted by the blue boxes. By comparison, FADRA produces sharper facial details, more faithful hair structures, and more stable appearances across frames. Although frame-based methods such as DiffBIR and RestoreFormer++ may restore plausible details in individual frames, they often suffer from temporal flickering and structural inconsistency. In contrast, FADRA maintains better identity stability and temporal coherence throughout the sequence. 
Notably, all results on the CelebV-HQ test set are obtained without any dataset-specific fine-tuning or hyper-parameter adjustment, demonstrating the strong cross-dataset generalization capability of our method.

\vspace{-2mm}

\subsection{Real-world Evaluation}

\begin{figure*}[h!]
    \centering
    \includegraphics[width=1\textwidth]{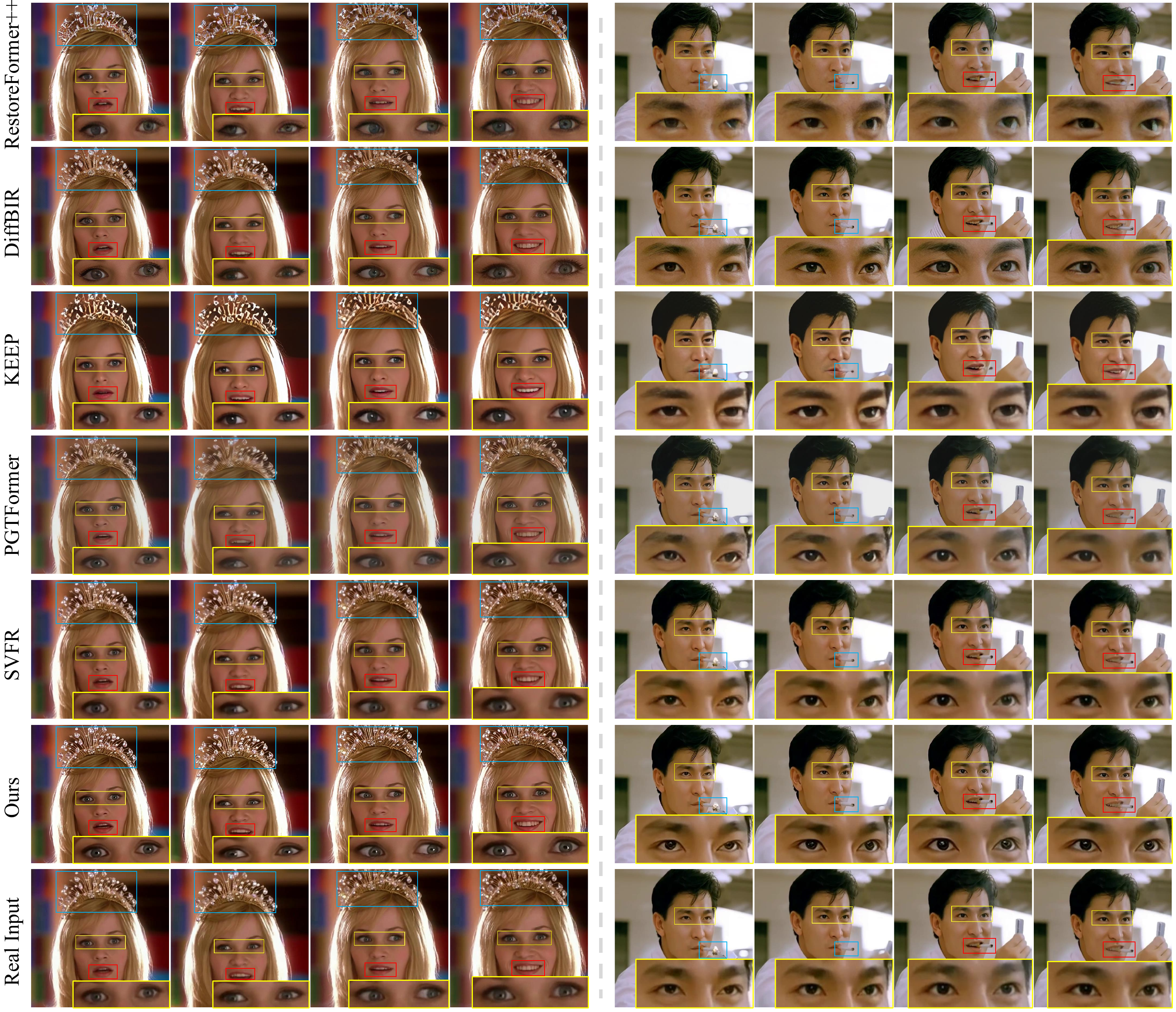} 
\caption{Visual comparison of restoration results from different methods on real movies. For two image-based methods, RestoreFormer++ and DiffBIR, we apply them frame-by-frame to the degraded videos. To facilitate a clearer comparison of local details, we highlight the eye region with yellow boxes, the mouth and teeth region with red boxes, and other non-facial regions, such as the crown and the cigarette, with blue boxes. }
    \label{fig:visual_compare_real-2}
    \vspace{-3mm}
\end{figure*}

To further demonstrate the practical utility and robustness of FADRA, we evaluate our model on challenging real-world movie sequences. Unlike synthetic datasets, these videos contain real-world degradation, severe compression, challenging lighting, and motion blur that are often out-of-distribution for standard restoration models.

\begin{table}[t]
\centering
\caption{User study on real-world videos. Quality, IDC, and TC are rated on a 1--5 scale.}
\label{tab:user}
\resizebox{0.49\textwidth}{!}{
\begin{tabular}{lcccc}
\toprule
\textbf{Method} 
& \textbf{Quality} $\uparrow$ 
& \textbf{IDC} $\uparrow$
& \textbf{TC} $\uparrow$
& \textbf{Best Vote \%} $\uparrow$\\
\midrule
RestoreFormer++    & 2.12 & 2.25 & 2.18 & 3.81\\
DiffBIR & 3.00 & 3.02 & 2.89 & 8.57\\
KEEP    & 2.17 & 1.74 & 2.02 & 2.38\\
SVFR    & \underline{3.35} & \underline{3.67} & \underline{3.60} & \underline{16.19}\\
\rowcolor{myblue2!30} 
\textbf{Ours} 
        & \textbf{4.36} & \textbf{4.32} & \textbf{4.31} & \textbf{69.05}\\
\bottomrule
\end{tabular}
}
\vspace{-3mm}
\end{table}

\begin{table}[t]
\centering
\caption{The ablation study of each component of the proposed method on the VFHQ test set. The Baseline refers to the model without the proposed FAL and RRAH. }
\label{tab:component}
\begin{tabular}{ccccccc}
\toprule

\textbf{FAL}  &\textbf{RRAH}
& \textbf{PSNR} $\uparrow$ 
& \textbf{SSIM} $\uparrow$ 
& \textbf{LPIPS} $\downarrow$ 
& \textbf{IDD} $\downarrow$
& \textbf{FVD} $\downarrow$\\
\midrule

\xmark 
& \xmark 
& 27.52
& 0.8249
& 0.2581
& 0.3046
& 64.34 \\

\cmark
&\xmark
& 28.56
& 0.8368
& 0.2521
& 0.2944
& 58.32 \\

\xmark
&\cmark
& 29.67
& 0.8550
& 0.2402
& \textbf{0.2806}
& 40.69 \\

\rowcolor{myblue2!30} 
\cmark
&\cmark
& \textbf{29.95}
& \textbf{0.8576}
& \textbf{0.2378}
& 0.2912
& \textbf{38.97} \\

\bottomrule
\end{tabular}
\vspace{-3mm}
\end{table}

\subsubsection{User Study}
To quantitatively assess the perceptual quality of the restored results on real-world videos, we conduct a user study on 10 real-world video sequences and collect 21 valid responses. As reported in Table~\ref{tab:user}, FADRA achieves the highest scores in visual quality, identity consistency (IDC), temporal consistency (TC), and best-vote preference, indicating its superior perceptual fidelity and temporal robustness in practical scenarios.

\subsubsection{Qualitative Comparison}
Qualitative comparisons are shown in Fig.~\ref{fig:visual_compare_real-2}. Existing methods often produce blurry facial details, unnatural tooth textures, ringing artifacts, or identity drift under complex real-world degradations. In particular, occlusions around the mouth region and rapid expression changes further challenge the ability of competing methods to preserve facial structure and temporal stability. In contrast, FADRA restores clearer eye and mouth details, better preserves identity, and maintains more stable facial geometry across frames. It also reconstructs non-facial details, such as accessories, with fewer temporal fluctuations. These results demonstrate that, although trained with synthetic degradations, FADRA generalizes well to in-the-wild video content and provides visually faithful, identity-preserving, and temporally coherent restorations on challenging movie sequences.

\vspace{-4mm}
\subsection{Ablation Studies}

\subsubsection{Component Effect Analysis}

We conduct ablation studies to analyze the contribution of each core component in FADRA. As shown in Table~\ref{tab:component}, introducing FAL to train the baseline in the second row brings consistent improvements across all five metrics. This improvement demonstrates that by providing frequency-domain supervision with HVS-inspired reweighting, the FAL module effectively enhances the reconstruction of fine-grained facial details. Comparing the third row with the baseline, the RRAH alone brings more substantial gains, particularly in structural fidelity and identity preservation, where PSNR/SSIM are boosted from 27.52/0.825 to 29.67/0.855, while LPIPS/IDD are reduced from 0.258/0.3 to 0.24/0.28. Furthermore, RRAH significantly improves temporal coherence, with the FVD score reduced from 64.34 to 40.69. 
These results indicate that the RRAH  with the LQ-guided ``re-examination'' mechanism is crucial for better capturing fine facial details while maintaining the inherent temporal priors of the pre-trained model. When combining both FAL and RRAH, our full model attains the best overall performance. Although the IDD score is slightly higher than that of the RRAH-only variant, it still surpasses the baseline and FAL-only models, while delivering the strongest frame-level and video-level quality. Overall, these ablations confirm that FAL and RRAH are complementary and jointly enable FADRA to produce high-fidelity and temporally coherent results.

\begin{table*}[t]
\centering
\caption{The ablation study of the Repeated Residual Adaptation Head (RRAH) on the VFHQ test set. ``$v \& z_{lq}$'' denotes that the input is the concatenation of the initial velocity prediction $v$ and the original low-quality latent $z_{lq}$. The Baseline corresponds to the model without FAL and RRAH. The second and third rows introduce the RRAH implemented with two Conv3D layers. The last two rows replace Conv3D with a single DiT block as the adaptation head. Here, 16 is the number of latent channels, $F$ denotes the temporal latent length, and 64 $\times$ 64 is the spatial latent resolution.
}

\label{tab:RRAH_ablation}
\resizebox{\textwidth}{!}{\begin{tabular}{l|ccccccccc}
\toprule
\textbf{RRAH Variation} 
& \textbf{Input} 
& \textbf{Input Size} 
& \textbf{Output Size} 
& \textbf{Layer} 
& \textbf{PSNR} $\uparrow$ 
& \textbf{SSIM} $\uparrow$ 
& \textbf{LPIPS} $\downarrow$ 
& \textbf{IDD} $\downarrow$
& \textbf{FVD} $\downarrow$\\
\midrule
Baseline
& -- 
& -- 
& -- 
& -- 
& 27.52
& 0.8249
& 0.2581
& 0.3046 
& 64.34 \\

Conv3D
& $v$ 
& 16$\times$$F$$\times$64$\times$64
& 16$\times$$F$$\times$64$\times$64
& 2
& 28.48
& 0.8538
& 0.2376
& 0.2980
& 64.64 \\

Conv3D
& $v$ \& ${z}_{lq}$ (Concat)
&  32$\times$$F$$\times$64$\times$64
&  16$\times$$F$$\times$64$\times$64
& 2
& 29.06
& 0.8540
& 0.2386
& 0.2832
& 56.96 \\

DiT Block
& $v$
& 16$\times$$F$$\times$64$\times$64
& 16$\times$$F$$\times$64$\times$64
& 1 
& 28.89
& \textbf{0.8575}
& \textbf{0.2354}
& 0.2863
& 48.88 \\

\rowcolor{myblue2!30} 
DiT Block
& $v$ \& ${z}_{lq}$ (Concat)
& 32$\times$$F$$\times$64$\times$64
& 16$\times$$F$$\times$64$\times$64
& 1 
& \textbf{29.67}
& 0.8550
& 0.2402
& \textbf{0.2806}
& \textbf{40.69} \\

\bottomrule
\end{tabular}
}
\vspace{-4mm}
\end{table*}

\begin{table}[t]
\centering
\caption{Ablation study on the complementary effect of LoRA and RRAH in FADRA on the VFHQ test set. Without LoRA, the newly added Conv3D parameters for LQ fusion are trained; with LoRA, the Conv3D is adapted through LoRA parameters.}
\label{tab:lora_rrah}
\begin{tabular}{ccccccc}
\toprule

\textbf{LoRA}  &\textbf{RRAH}
& \textbf{PSNR}$\uparrow$ 
& \textbf{SSIM}$\uparrow$ 
& \textbf{LPIPS}$\downarrow$ 
& \textbf{IDD}$\downarrow$
& \textbf{FVD}$\downarrow$\\
\midrule

\xmark 
& \xmark 
& 13.05	
& 0.4706	
& 0.5492	
& 1.0469
& 2389.42 \\

\cmark 
& \xmark 
& \underline{27.52}
& \underline{0.8249}
& \underline{0.2581}
& \underline{0.3046}
& \underline{64.34}  \\

\xmark 
& \cmark 
& 24.16	
& 0.7561	
& 0.3608	
& 0.6408
& 362.84 \\

\rowcolor{myblue2!30}
\cmark 
& \cmark 
& \textbf{29.67}
& \textbf{0.8550}
& \textbf{0.2402}
& \textbf{0.2806}
& \textbf{40.69} \\

\bottomrule
\end{tabular}
\vspace{-4mm}
\end{table}

\subsubsection{Ablation on Repeated Residual Adaptation Head}

We further investigate the structural design and the different combinations of input features of the RRAH to determine the optimal mechanism for predicting a residual update at each flow-matching velocity prediction step. As summarized in Table~\ref{tab:RRAH_ablation},  compared with the baseline, all RRAH variants bring clear performance gains, confirming the effectiveness of residual adaptation. By comparing the second and third rows, as well as the fourth and fifth rows, we observe that concatenating the predicted velocity $v$ with the LQ latent condition ${z}_{lq}$ consistently leads to significant performance gains. This demonstrates that explicit guidance from the LQ latent is essential for the adaptation head to accurately capture structural residuals and maintain identity consistency. By comparing the second row with the fourth, and the third row with the fifth, it can be seen that replacing Conv3D with a DiT block yields even larger improvements. The full configuration, where the DiT block operates on the concatenation of $v$ and ${z}_{lq}$, delivers the best overall performance. These results show that a transformer-based residual adaptation head with joint velocity–latent conditioning is crucial for capturing both spatial details and temporal dynamics, enabling FADRA to produce high-fidelity and temporally coherent restoration results.

We also examine the complementary role of LoRA and RRAH. As shown in Table~\ref{tab:lora_rrah}, introducing either LoRA or RRAH improves the baseline performance, indicating that both modules contribute to adapting the frozen T2V prior to VFR. When used together, they achieve the best results, showing that backbone-level LoRA adaptation and LQ-guided residual refinement are complementary for improving restoration fidelity and temporal coherence.

\begin{figure*}[t]
    \centering
    \includegraphics[width=1\textwidth]{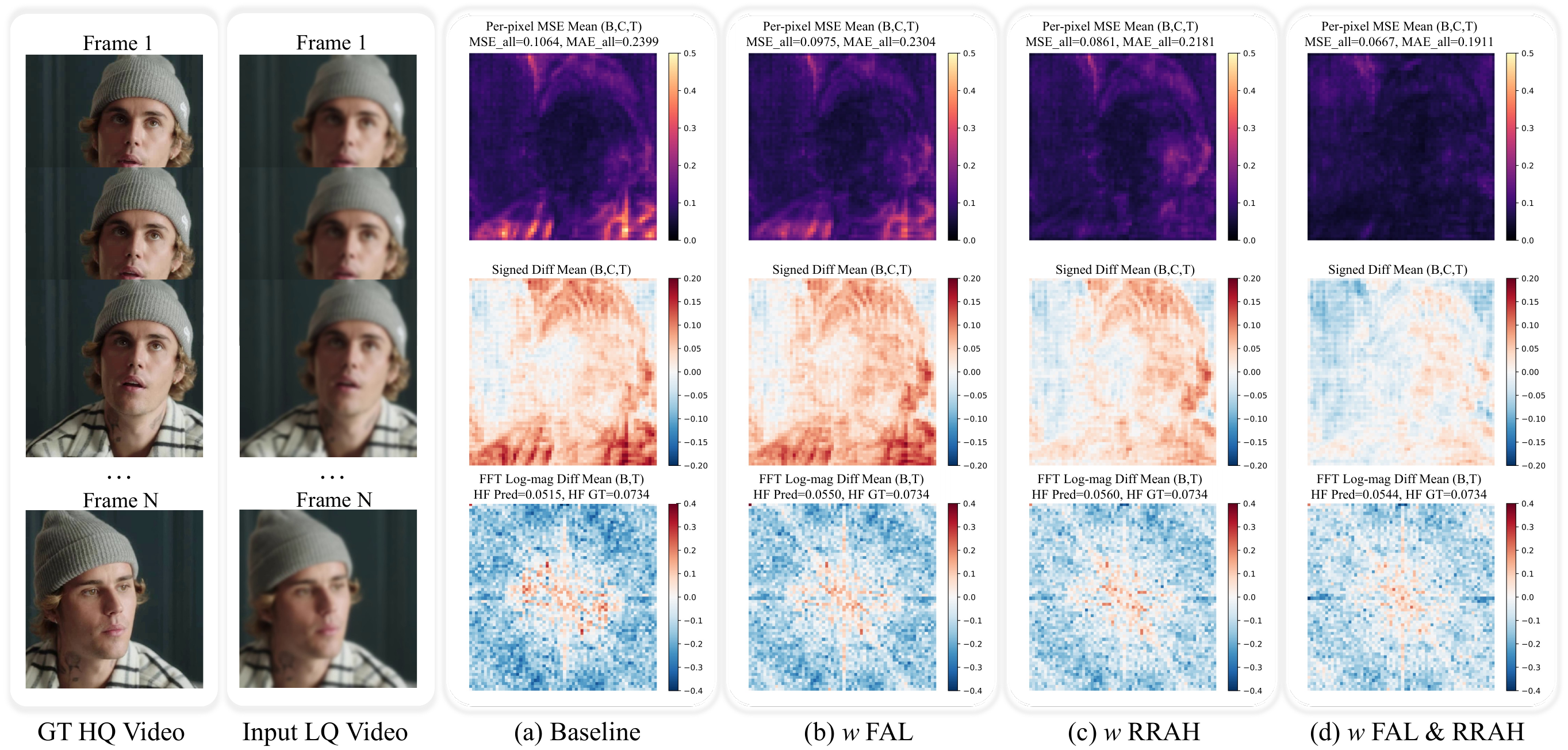} 
    \caption{Visualization of latent-space error and frequency statistics between ground-truth and predicted latents. MSE: mean squared error; MAE: mean absolute error; Signed Diff: signed difference; HF: high-frequency energy ratio; FFT Log-Mag Diff: FFT log-magnitude difference.}
    \label{fig:latent_error}
    \vspace{-3mm}
\end{figure*}

\subsubsection{Ablation on Frequency-Aware Loss} 
\begin{table}[t]
\centering
\caption{The ablation study of Frequency-Aware Loss (FAL) on the VFHQ test set.}
\label{tab:freq_loss_ablation}
\begin{tabular}{l|c}
\toprule
\textbf{Loss Design} / \textbf{Domain}  
& \textbf{PSNR}$\uparrow$ / \textbf{SSIM}$\uparrow$ / \textbf{LPIPS}$\downarrow$ / \textbf{IDD}$\downarrow$ / \textbf{FVD}$\downarrow$\\
\midrule
Baseline / Spatial 
& 27.52 / 0.8249 / 0.2581 / 0.3046 / 64.34 \\

FAL on $v$ / Frequency 
& 27.45 / \textbf{0.8400} / \textbf{0.2414} / 0.3430 / 72.86 \\

\rowcolor{myblue2!30} 
\textbf{FAL on $z$} / Frequency 
& \textbf{28.56} / 0.8368 / 0.2521 / \textbf{0.2944} / \textbf{58.32} \\

\bottomrule
\end{tabular}
\vspace{-5mm}
\end{table}

We also investigate the optimal design of the FAL by applying it to different optimization targets within the flow-matching framework. Specifically, we compare two strategies: applying FAL directly to the velocity $v$ versus applying FAL to the corresponding latent $z$. As shown in Table~\ref{tab:freq_loss_ablation}, applying FAL to the latent $z$ yields significantly better results than applying it to velocity $v$, with PSNR improving from 27.45 to 28.56, IDD decreasing from 0.3430 to 0.2944, and FVD decreasing from 72.86 to 58.32. This performance gap aligns with the physical intuition of frequency-domain supervision. Compared with velocity $v$, which is the difference between LQ and HQ latents, the latent $z$ directly corresponds to the visual content and structural information of the face. Since frequency-aware constraints are designed to supervise structural textures and perceptually important details, applying them in the latent space provides more direct and meaningful guidance for high-fidelity restoration. Consequently, we adopt the latent $z$ as the supervision domain for FAL in our final model, leading to superior identity preservation and temporal stability.

\subsubsection{Latent-Space Error and Frequency Analysis} 

To further investigate the internal mechanism of the proposed components, we conduct a fine-grained analysis of the discrepancy between the ground-truth latent encoded by the VAE and the predicted latent under different ablation settings. As illustrated in Fig.~\ref{fig:latent_error},  the top row shows pixel-wise reconstruction errors, the middle row shows the signed difference mean, and the bottom row depicts high-frequency statistics, including the high-frequency energy ratio and the FFT log-magnitude difference. From left to right, we compare the baseline, the model with FAL, the model with RRAH, and the full model with both FAL and RRAH. Compared with the baseline, incorporating either FAL or RRAH consistently reduces restoration discrepancies, as evidenced by the suppressed error magnitude, more neutral signed-difference maps, and high-frequency energy that is better aligned with the ground-truth latent. These visual improvements indicate that FAL effectively supervises the recovery of fine-grained textures and high-frequency details that are typically lost in the restoration process. Furthermore, the LQ-guided repeated residual refinement in RRAH encourages the model to more fully exploit degraded cues, thereby better capturing subtle facial details. When combining FAL and RRAH, the error maps become more uniform, demonstrating that the two modules are complementary and jointly lead to more accurate latent representations. These latent-space behaviors are consistent with the quantitative gains reported in Table~\ref{tab:component}.

\subsubsection{Analysis of Model Robustness to Pose Variations}
\begin{table}[t]
\centering
\caption{Analysis of model robustness to pose variations based on yaw-span intervals on the VFHQ test set.}
\label{tab:span_pose_compare}
\begin{tabular}{lcl}
\toprule
\textbf{Method} 
& \textbf{Interval}  
& \textbf{PSNR}$\uparrow$ / \textbf{SSIM}$\uparrow$ / \textbf{LPIPS}$\downarrow$ / \textbf{IDD}$\downarrow$ / \textbf{FVD}$\downarrow$\\
\midrule
\multirow{4}{*}{Ours}
& [0,15]   & 29.43 / 0.8447 / 0.2454 / 0.2907 / 37.48 \\
& [15,30]  & 30.27 / 0.8616 / 0.2329 / 0.2867 / 37.27 \\
& [30,90]   & 30.43 / 0.8808 / 0.2304 / 0.3046 / 47.31 \\

\rowcolor{myblue2!30} 
& Overall     & 29.95 / 0.8576 / 0.2378 / 0.2912 / 38.97 \\
\midrule
\multirow{4}{*}{PGTFormer}
& [0,15]   & 25.47 / 0.7871 / 0.2632 / 0.3781 / 61.57 \\
& [15,30]  & 25.85 / 0.7784 / 0.2607 / 0.3978 / 76.12 \\
& [30,90]   & 26.11 / 0.7985 / 0.2557 / 0.4019 / 98.09 \\
\rowcolor{myblue2!30} 
& Overall     & 25.73 / 0.7853 / 0.2610 / 0.3902 / 73.52 \\
\midrule
\multirow{4}{*}{DiffBIR}
& [0,15]   & 27.14 / 0.7574 / 0.2860 / 0.3578 / 49.97 \\
& [15,30]  & 27.70 / 0.7705 / 0.2832 / 0.3557 / 55.89 \\
& [30,90]   & 28.26 / 0.7878 / 0.2881 / 0.3577 / 55.53 \\
\rowcolor{myblue2!30} 
& Overall     & 27.56 / 0.7678 / 0.2852 / 0.3569 / 53.35 \\
\bottomrule
\end{tabular}
\vspace{-3mm}
\end{table}

To evaluate the robustness of our model to head pose changes, we further conduct a pose-aware analysis on the VFHQ test set. Specifically, we utilize SynergyNet~\cite{SynergyNet} to estimate the head pose of every frame in each video. Then, we calculate the difference between the maximum and minimum yaw angles for each video as the pose span. According to this, videos are grouped into three intervals, corresponding to increasing pose variation. As reported in Table~\ref{tab:span_pose_compare}, we select DiffBIR and PGTFormer as representative diffusion-based image restoration and Transformer-based video restoration methods, respectively. When the yaw span increases, all compared methods experience a noticeable decline in performance, particularly in identity preservation and temporal coherence, as indicated by the rising IDD and FVD scores. However, our FADRA consistently outperforms PGTFormer and DiffBIR across all intervals, demonstrating superior robustness. In the most challenging range of $[30^\circ, 90^\circ]$, FADRA maintains the best FVD score of 47.31, while PGTFormer and DiffBIR show higher FVD scores of 98.09 and 55.53, respectively. These results demonstrate that FADRA is more robust to large head pose variations than existing image-based and video-based face restoration approaches.

\vspace{-3mm}

\subsection{Parameter counts and computational overhead}

We further analyze the parameter counts and computational overhead of FADRA. As reported in Table~\ref{tab:Parameters}, FADRA only introduces a small number of trainable parameters through three lightweight components, including the LoRA adapters, the spatio-temporal feature fusion module, and the Repeated Residual Adaptation Head (RRAH). Compared with the large frozen generative backbone, these additional parameters are modest, indicating that FADRA does not rely on full-model fine-tuning to achieve high-quality video face restoration. Instead, it adapts the pretrained diffusion prior in a parameter-efficient manner, where the frozen backbone provides strong generative and temporal modeling capabilities, while the newly introduced modules focus on degradation-aware feature fusion and residual refinement.
\begin{table}[t]
\centering
\caption{Parameter counts of the frozen backbone and trainable adaptation modules.}
\label{tab:Parameters}
\resizebox{0.495\textwidth}{!}{
\begin{tabular}{lcccc}
\toprule
\textbf{Part} & \textbf{Wan} & \textbf{LoRA} & \textbf{Feature Fusion} & \textbf{RRAH} \\
\midrule
\textbf{Parameter (M)} & 1419.0 & 94.3 & 0.2 & 72.3 \\
\bottomrule
\end{tabular}
}
\vspace{-3mm}
\end{table}

\begin{table}[t]
\centering
\caption{Computational overhead introduced by RRAH. FPS and processing time are measured under the same inference setting as Table~\ref{tab:FPS_compare}.}
\label{tab:Cost}
\begin{tabular*}{0.492\textwidth}{@{\extracolsep{\fill}}lcc@{\extracolsep{\fill}}}
\toprule
\textbf{Method} & \textbf{FPS $\uparrow$}  & \textbf{Inference Time (s) $\downarrow$}\\
\midrule
 \textbf{w/o RRAH} & \textbf{0.885} &  \textbf{182} \\
 \textbf{Ours} & \underline{0.866} & \underline{186}\\
\bottomrule
\end{tabular*}
\vspace{-4mm}
\end{table}

The computational cost of FADRA is also limited. Although RRAH performs step-wise residual adaptation during the refinement process, its design is lightweight and only introduces marginal extra latency. Under the same evaluation setting as Table~\ref{tab:FPS_compare}, the latency comparison in Table~\ref{tab:Cost} shows that RRAH adds only 2.2\% latency, while the ablation results in Table~\ref{tab:component} show that it reduces the baseline FVD by 36.8\%, demonstrating that the proposed refinement mechanism brings significant temporal and perceptual improvements with negligible additional computational burden. These results suggest that FADRA achieves a favorable trade-off between restoration quality and efficiency.

\section{Conclusion}

In this paper, we present FADRA, a frequency-aware diffusion framework with iterative residual adaptation designed to address the challenges of spatial fidelity and temporal coherence in video face restoration. FADRA leverages a pre-trained text-to-video diffusion model to exploit its strong generative and temporal priors, while effectively injecting degraded video information, making the framework well-suited for VFR. Specifically, RRAH performs repeated residual adaptation that iteratively refines the predicted velocity using low-quality cues, effectively capturing fine facial details without compromising temporal stability. Furthermore, the introduction of a Frequency-Aware Loss (FAL) provides explicit supervision in the spectral domain, ensuring the structural integrity of perceptually important facial details. Extensive quantitative and qualitative evaluations on synthetic and real-world degraded videos demonstrate that FADRA shows clear advantages in identity preservation, temporal consistency, and perceptual quality, while effectively suppressing flickering artifacts across frames. These results validate the effectiveness of adapting video diffusion priors for high-quality and temporally coherent video face restoration. Future work will focus on improving the efficiency and robustness of FADRA toward more practical real-world VFR, especially under large pose changes, complex motions, and diverse in-the-wild degradations.

\bibliographystyle{IEEEtran}
\bibliography{main}

\vspace{11pt}

\vfill

\end{document}